\documentclass[11pt,a4paper]{article}
\usepackage[hyperref]{acl2017}
\usepackage{times}
\usepackage{latexsym}
\usepackage{graphicx}
\usepackage{amsmath}
\usepackage{amsfonts}
\usepackage{graphicx}
\usepackage{color}
\usepackage{url}
\usepackage{algorithm}
\usepackage[noend]{algpseudocode}
\usepackage{enumitem}
\usepackage[font={footnotesize}]{caption}

\aclfinalcopy 
\def\aclpaperid{606} 

\title{Neural Symbolic Machines: \\ 
 Learning Semantic Parsers on Freebase with Weak Supervision}

\author{ 
  \bf{Chen Liang}\thanks{\hspace{1mm} Work done while the author was interning at Google},
  \bf{Jonathan Berant}\thanks{\hspace{1mm} Work done while the author was   visiting Google},
  \bf{Quoc Le},
  \bf{Kenneth D. Forbus},
  \bf{Ni Lao} \\
Northwestern University, Evanston, IL \\
Tel-Aviv University, Tel Aviv-Yafo, Israel \\
Google Inc., Mountain View, CA \\
\{chenliang2013,forbus\}@u.northwestern.edu,
joberant@cs.tau.ac.il,
\{qvl,nlao\}@google.com
}

\date{}

\begin{document}
\maketitle
\begin{abstract}



Harnessing the statistical power of neural networks to perform language understanding and symbolic reasoning is difficult, when it requires executing
efficient discrete 
operations against a large knowledge-base.
%
%
%
In this work, we introduce a \textit{Neural Symbolic Machine} (NSM), which contains 
(a) a neural ``programmer", i.e., a sequence-to-sequence model that maps language utterances to programs and utilizes a \textit{key-variable memory} to handle compositionality 
(b) a symbolic ``computer", i.e., a Lisp interpreter that performs program execution, and helps find good programs by pruning the search space. 
%
We apply REINFORCE to directly optimize the task reward of this structured prediction problem.
To train with weak supervision and improve the stability of REINFORCE 
we augment it with an  \textit{iterative maximum-likelihood} training process. 
NSM outperforms the state-of-the-art on the \textsc{WebQuestionsSP} dataset when trained from question-answer pairs only, without requiring any feature engineering or domain-specific knowledge.

\end{abstract}

\section{Introduction}

\begin{figure}[t]
\centering
\includegraphics[width=2.5in]{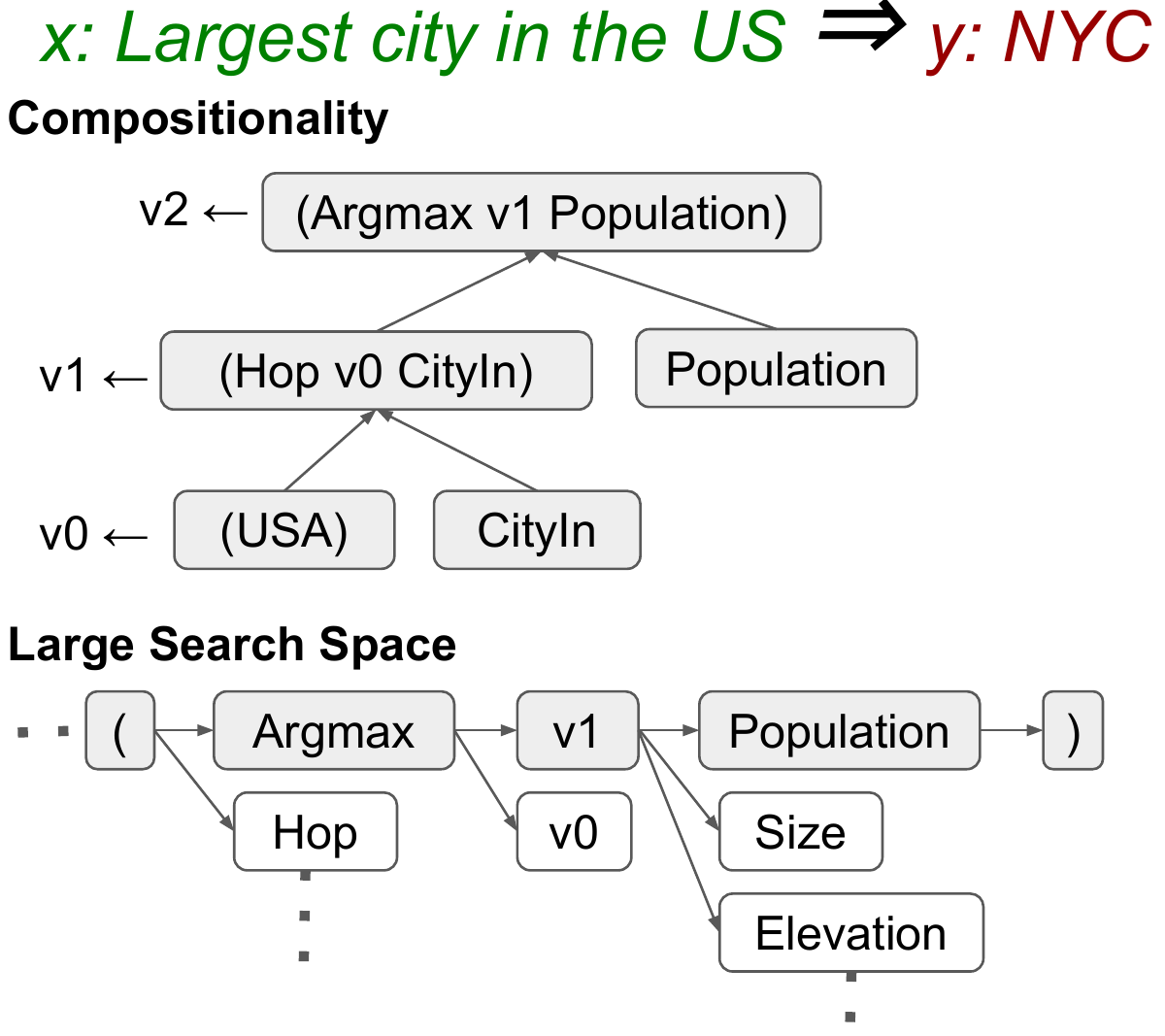}
\vspace{-0.1in}
\caption{\label{fig:challenges} The main challenges of training a semantic parser from weak supervision: (a) \emph{compositionality}: we use variables ($v_0, v_1, v_2$) to store execution results of intermediate generated programs. (b) \emph{search}: we prune the search space and augment REINFORCE with pseudo-gold programs.
}
\end{figure}

Deep neural networks have achieved impressive performance in supervised classification and structured prediction tasks such as speech recognition \cite{hinton2012deep}, machine translation \cite{bahdanau2014align,wu2016gnmt} and more.
However, training neural networks for semantic parsing \cite{zelle96geoquery,zettlemoyer05ccg,liang11dcs} or program induction, where language is mapped to a symbolic representation that is executed by an executor, through weak supervision remains challenging. This is because the model must interact with a symbolic executor through non-differentiable operations 
to search over a large program space.

In semantic parsing, recent work handled this \cite{dong2016language,jia2016data} by training from manually annotated programs and avoiding program execution at training time. However, annotating programs is known to be expensive and scales poorly.
%
%
%
%
%
In program induction, attempts to address this problem  \cite{graves2014neural,reed2015neural,kaiser2015neural,graves2016hybrid,andreas2016compose}
either utilized low-level memory~\cite{zaremba2015reinforcement}, or required memory to be differentiable~\cite{Neelakantan2015NeuralPI,yin2015neural} so that the model can be trained with backpropagation.
This makes it difficult to use the efficient discrete operations and memory of a traditional computer, and limited the application to synthetic or small knowledge bases.

In this paper, we propose to utilize the memory and discrete operations of a traditional computer in a novel Manager-Programmer-Computer (MPC) framework for neural program induction, which integrates three components:
\begin{enumerate}[noitemsep,topsep=0pt]
\item A \textbf{``manager"} that provides weak supervision (e.g., \textsc{`NYC'} in Figure~\ref{fig:challenges}) through a reward indicating how well a task is accomplished. Unlike full supervision, weak supervision is easy to obtain at scale (Section \ref{sec:webquestions}).
\item A \textbf{``programmer"} that takes natural language as input and generates a program that is a sequence of tokens (Figure~\ref{fig:parsing}). 
The programmer learns from the reward and must overcome the hard search problem of finding correct programs (Section \ref{subsec:model}).
\item A \textbf{``computer"} that executes programs in a high level programming language. 
Its non-differentiable 
memory enables \textit{abstract}, \textit{scalable} and \textit{precise} operations, but makes training more challenging (Section \ref{subsec:training}). 
To help the ``programmer" prune the search space, it provides a friendly \textit{neural computer interface},
which detects and eliminates invalid choices (Section \ref{subsec:language}).
\end{enumerate}

Within this framework, we introduce the Neural Symbolic Machine (NSM) and apply it to semantic parsing. NSM contains a neural sequence-to-sequence (seq2seq) ``programmer" \cite{sutskever2014sequence} and a symbolic non-differentiable Lisp interpreter (``computer") that executes programs against a large knowledge-base (KB).

Our technical contribution in this work is threefold. First, to support language \textit{compositionality}, we augment the standard seq2seq model with a \textit{key-variable memory} to save and reuse intermediate execution results (Figure~\ref{fig:challenges}). 
This is a novel application of pointer networks ~\cite{vinyals2015pointer} to compositional semantics.

Second, to alleviate the search problem of finding correct programs when training from question-answer pairs,
we use the computer to execute partial programs and prune the programmer's search space by checking the syntax and semantics of generated programs.
This generalizes 
the weakly supervised semantic parsing framework~\cite{liang11dcs,berant2013semantic} 
by leveraging semantic denotations during structural search.
 
Third, to train from weak supervision and directly maximize the expected reward
we turn to the REINFORCE~\cite{Williams92simplestatistical} algorithm.
Since learning from scratch is difficult for REINFORCE,
we combine it with an \textit{iterative maximum likelihood} (ML) training process, where beam search is used to find pseudo-gold programs,
which are then used to augment the objective of REINFORCE. 


On the \textsc{WebQuestionsSP} dataset \cite{yih2016webquestionssp}, NSM achieves new state-of-the-art results with weak supervision, significantly closing the gap between weak and full supervision for this task. Unlike prior works, it is trained end-to-end, and does not require feature engineering or domain-specific knowledge. 


\section{Neural Symbolic Machines}
We now introduce NSM by 
first describing the ``computer", a non-differentiable Lisp interpreter that executes programs against a large KB and provides code assistance (Section \ref{subsec:language}). We then propose a seq2seq model (``programmer") that supports compositionality using a key-variable memory to save and reuse intermediate results (Section~\ref{subsec:model}).
Finally, we describe a training procedure that is based on REINFORCE, but is augmented with pseudo-gold programs found by an iterative ML training  procedure (Section~\ref{subsec:training}).

Before diving into details, we define the \textit{semantic parsing} task:
given a knowledge base $\mathbb{K}$, and a question $x=(w_1, w_2, ..., w_m)$, produce a program or logical form $z$ that when executed against $\mathbb{K}$ generates the right answer $y$.
Let $\mathcal{E}$ denote a set of entities (e.g.,  \textsc{AbeLincoln}),\footnote{We  also consider numbers (e.g., ``1.33'') and date-times (e.g., ``1999-1-1'') as entities.} and let $\mathcal{P}$ denote a set of properties (e.g., \textsc{PlaceOfBirth}). A knowledge base $\mathbb{K}$ is a set of assertions or triples $(e_1,p,e_2) \in \mathcal{E} \times \mathcal{P} \times \mathcal{E}$, such as  (\textsc{AbeLincoln}, \textsc{PlaceOfBirth}, \textsc{Hodgenville}). 

\subsection{Computer: Lisp Interpreter with Code Assistance}
\label{subsec:language}

Semantic parsing typically requires using a set of operations to query the knowledge base and process the results. 
Operations learned with neural networks such as addition and sorting do not perfectly generalize to inputs that are larger than the ones observed in the training data~\cite{graves2014neural,reed2015neural}. 
In contrast, operations implemented in high level programming languages are \textit{abstract}, \textit{scalable}, and \textit{precise}, thus generalizes perfectly to inputs of arbitrary size.
Based on this observation, we implement operations necessary for semantic parsing with an ordinary
programming language instead of trying to learn them with a neural network. 

\begin{table*}[ht]
\centering
  \begin{tabular}{ cl}
  	\hline
    $($ \textit{Hop} $r$ $p$ $)$ $\Rightarrow$ $\{e_2 | e_1 \in r, (e_1, p, e_2) \in \mathbb{K}\}  $\\ 
    $($ \textit{ArgMax}  $r$ $p$ $)$ $\Rightarrow$ $\{e_1 | e_1 \in r, \exists e_2 \in \mathcal{E}: (e_1, p, e_2) \in \mathbb{K}, \forall e : (e_1, p, e) \in \mathbb{K}, e_2 \geq e \}$\\ 
    $($ \textit{ArgMin}  $r$ $p$ $)$ $\Rightarrow$ $\{e_1 | e_1 \in r, \exists e_2 \in \mathcal{E}: (e_1, p, e_2) \in \mathbb{K}, \forall e : (e_1, p, e) \in \mathbb{K}, e_2 \leq e \}$\\ 
    $($ \textit{Filter}  $r_1$ $r_2$ $p$  $)$ $\Rightarrow$ $\{e_1 | e_1 \in r_1, \exists e_2 \in r_2:  (e_1, p, e_2) \in \mathbb{K}\}  $ \\ 
    \hline
  \end{tabular}
\caption{Interpreter functions. $r$ represents a variable, $p$ a property in Freebase.  $\geq$ and $\leq$ are defined on numbers and dates. 
}
\label{tab-functions}
\end{table*}

We adopt a Lisp interpreter 
as the ``computer".
A program $C$ is a list of expressions $(c_1 ... c_N)$, where
each expression is either a special token ``\textit{Return}" indicating the end of the program,
or a list of tokens enclosed by parentheses ``$( F A_1 ... A_K )$". $F$ is 
a function,
which takes as input $K$ arguments of specific types. Table~\ref{tab-functions} defines the semantics of each function and the types of its arguments (either a property $p$ or a variable $r$). When a function is executed, it returns an entity list that is the expression's denotation in $\mathbb{K}$, and save it to a new variable. 

By introducing variables that save the intermediate results of execution, the program naturally models \emph{language compositionality} and describes from left to right a bottom-up derivation of the full meaning of the natural language input, which is convenient in a seq2seq model (Figure~\ref{fig:challenges}). This is reminiscent of the floating parser \cite{wang2015building,Pasupat2015CompositionalSP}, where a derivation tree that is not grounded in the input is incrementally constructed.

The set of programs defined by our functions is equivalent to the subset of $\lambda$-calculus presented in \cite{yih2015semantic}.
We did not use full Lisp programming language here, because constructs like control flow and loops are unnecessary for most current semantic parsing tasks, and it is simple to add more functions to the model when necessary.

To create a friendly \textit{neural computer interface}, the interpreter provides code assistance to the programmer by producing a list of valid tokens at each step. 
First, a valid token should not cause a syntax error: e.g., if the previous token is ``$($", the next token  must be a function name, and if the previous token is ``\textit{Hop}", the next token must be a variable. 
More importantly, a valid token should not cause a semantic (run-time) error: this is detected using the denotation saved in the variables. 
For example, if the previously generated tokens were ``$($ \textit{Hop} $r$", the next available token is restricted to properties $\{p \mid \exists e,e': e \in r, (e, p, e') \in \mathbb{K}\}$ that are reachable from entities in $r$ in the KB. 
These checks are enabled by the variables and can be derived from the definition of the functions in Table \ref{tab-functions}.
The interpreter prunes the ``programmer"'s search space by orders of magnitude, and enables learning from weak supervision on a large KB.

\subsection{Programmer: Seq2seq Model with Key-Variable Memory}
\label{subsec:model}

\begin{figure*}[h!]
\centering
\includegraphics[width=\textwidth]{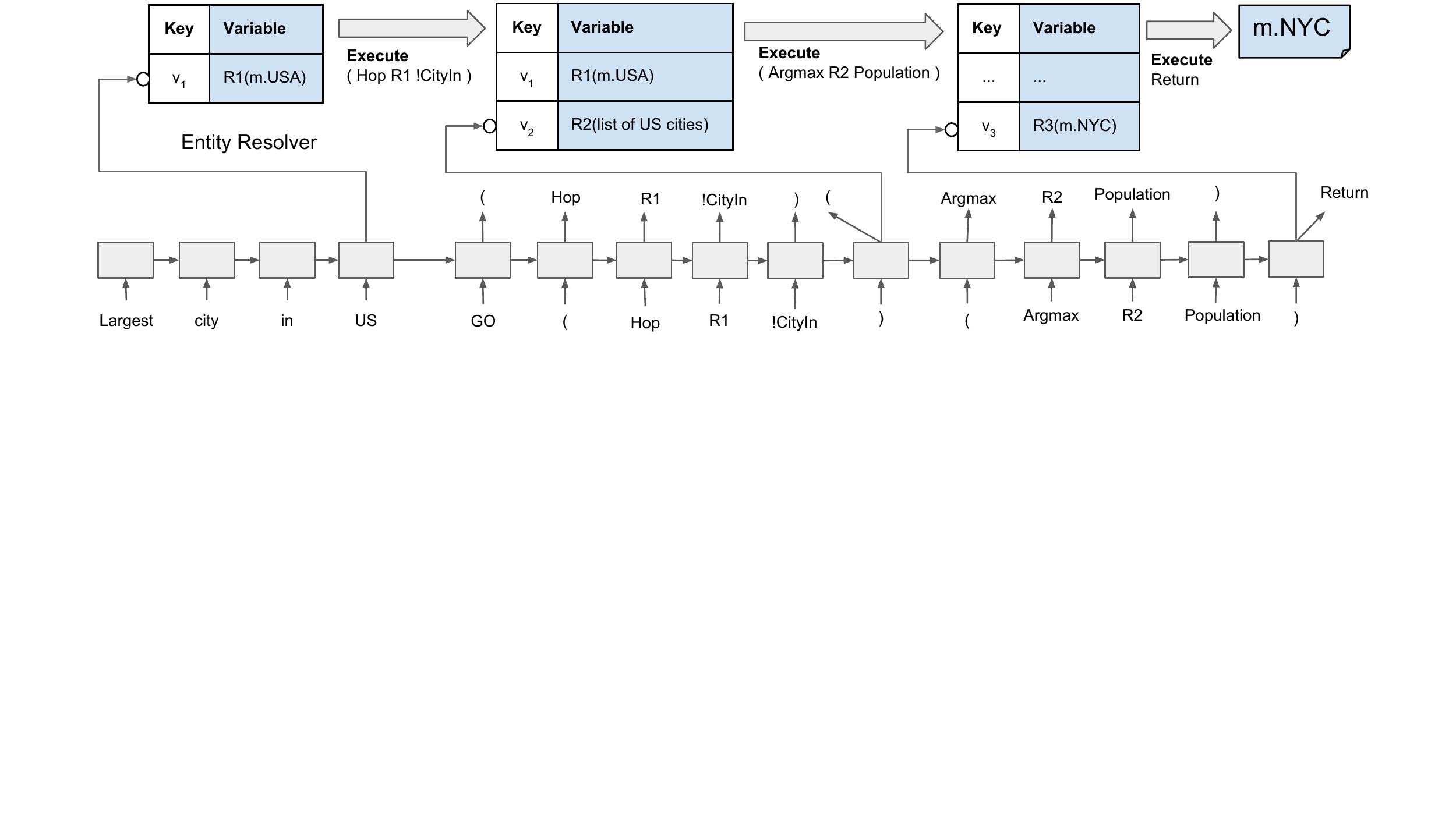}
\vspace{-2.4in}
\caption{\label{fig:parsing} Semantic Parsing with NSM. 
The key embeddings of the key-variable memory are the output of the sequence model at certain encoding or decoding steps.
For illustration purposes, we also show the values of the variables in parentheses, but the sequence model never sees these values, and only references them with the name of the variable (``$R_1$''). A special token ``\textit{GO}'' indicates the start of decoding, and ``\textit{Return}'' indicates the end of decoding.}
\end{figure*}

Given the ``computer'', the ``programmer'' needs to map natural language into a program, which is a sequence of tokens that reference operations and values in the ``computer". 
We base our programmer on a standard seq2seq model with attention, but extend it with a key-variable memory that allows the model to learn to represent and refer to program variables (Figure \ref{fig:parsing}).

Sequence-to-sequence models consist of two RNNs, an encoder and a decoder. We used a 1-layer GRU \cite{cho2014:gru}
for both the encoder and  decoder. 
Given a sequence of words ${w_1, w_2 ... w_m}$, each word $w_t$ is mapped to an embedding $q_t$ (embedding details are in Section~\ref{sec:experiment}). 
Then, the encoder reads these embeddings and updates its hidden state step by step using
$h_{t+1} = GRU(h_{t}, q_{t}, \theta_{Encoder})$, where $\theta_{Encoder}$ are the GRU parameters.
The decoder updates its hidden states $u_t$
by $u_{t+1} = GRU(u_{t}, c_{t-1}, \theta_{Decoder})$, where $c_{t-1}$ is the embedding of last step's output token $a_{t-1}$, and $\theta_{Decoder}$ are the GRU parameters.
The last hidden state of the encoder $h_{T}$ is used as the decoder's initial state. 
We also adopt a dot-product attention similar to \newcite{dong2016language}.
The tokens of the program ${a_1, a_2 ... a_n}$ are generated one by one using a softmax over the vocabulary of valid tokens at each step, as provided by the ``computer" (Section \ref{subsec:language}). 

To achieve compositionality, the decoder must learn to represent and refer to intermediate variables whose value was saved in the ``computer'' after execution.
Therefore, we augment the model with a \textbf{key-variable memory}, where
each entry has two components:  a continuous embedding key $v_i$, and a corresponding variable token $R_i$ referencing the value in the ``computer" (see Figure~\ref{fig:parsing}). 
During encoding, we use an entity linker to link text spans (e.g., \emph{``US"}) to KB entities.
For each linked entity we add a memory entry where the key is the average of GRU hidden states over the entity span, and the variable token ($R_1$) is the name of a variable in the computer holding the linked entity (\emph{m.USA}) as its value. 
During decoding, when a full expression is generated (i.e., the decoder generates ``$)$"), it gets executed, and the result is stored as the value of a new variable in the ``computer". This variable is keyed by the GRU hidden state at that step.
When a new variable $R_1$ with key embedding $v_1$ is added into the key-variable memory, the token $R_1$ is added into the decoder vocabulary with  $v_1$ as its embedding. 
The final answer returned by the ``programmer'' is the value of the last computed variable.

Similar to pointer networks ~\cite{vinyals2015pointer}, the key embeddings for variables are dynamically generated 
for each example.
During training, the model learns to represent variables by backpropagating gradients from a time step where a variable is selected by the decoder, through the key-variable memory, to an earlier time step when the key embedding was computed.
Thus, the encoder/decoder learns to generate representations for variables such that they can be used at the right time to construct the correct program. 

While the key embeddings are differentiable, the values referenced by the variables (lists of entities), 
stored in the ``computer'', are 
symbolic and non-differentiable. 
This distinguishes the key-variable memory from other memory-augmented neural networks that use continuous differentiable embeddings as the values of memory entries \cite{weston2014memory,graves2016ntm}. 

\subsection{Training NSM with Weak Supervision} 
\label{subsec:training}
NSM executes non-differentiable operations against a KB, and thus end-to-end backpropagation is not possible. Therefore, we base our training procedure on REINFORCE \cite{Williams92simplestatistical,norouzi2016}. 
When the reward signal is sparse and the search space is large, it is common to utilize some full supervision to pre-train REINFORCE \cite{silver2016mastering}.
To train from weak supervision, we suggest an iterative ML procedure for finding pseudo-gold programs that will bootstrap REINFORCE.


\paragraph{REINFORCE}
\label{reinforce}

We can formulate training as a reinforcement learning problem: given a question $x$, the state, action and reward at each time step $t \in \{0, 1, ..., T\}$ are $(s_t, a_t, r_t)$.
Since the environment is deterministic, the state is defined by the question $x$ and the action sequence: $s_t=(x, a_{0:t-1})$, where  $a_{0:t-1}=(a_0, ..., a_{t-1})$ is the history of actions at time $t$. 
A valid action at time $t$ is $a_t \in A(s_t)$, where $A(s_t)$ is the set of valid tokens given by the ``computer". Since each action corresponds to a token, the full history $a_{0:T}$ corresponds to a program. 
The reward $r_t=I[t=T] \cdot F_1(x, a_{0:T})$ is non-zero only at the last step of decoding, and is the $F_1$ score computed comparing the gold answer and the answer generated by executing the program $a_{0:T}$. Thus, the cumulative reward of a program $a_{0:T}$ is 
\begin{equation*}
R(x,a_{0:T})=\sum_t r_t = F_1(x, a_{0:T}).
\end{equation*}

The agent's decision making procedure at each time
is defined by a policy, $\pi_\theta(s, a) = P_\theta(a_t=a|x, a_{0:t-1})$, where $\theta$ are the model parameters. Since the environment is deterministic, the probability of generating a program $a_{0:T}$ is 
\begin{equation*}
P_\theta(a_{0:T}|x) = \prod_{t} P_\theta(a_t \mid x, a_{0:t-1}).
\end{equation*}

We can define our objective to be the expected cumulative reward and use policy gradient methods such as REINFORCE for training. The objective and gradient are:
\begin{equation*}
\begin{split}
J^{RL}(\theta) =& \sum_{x} \mathbb{E}_{P_\theta(a_{0:T} \mid x)}[R(x,a_{0:T})], \\
\nabla_{\theta}  J^{RL}(\theta) =&   \sum_{x} \sum_{a_{0:T}} P_\theta(a_{0:T} \mid x) \cdot [R(x,a_{0:T})-\\ 
& B(x)] \cdot \nabla_{\theta} \log P_\theta(a_{0:T} \mid x),
\end{split}
\end{equation*}
where $B(x)= \sum_{a_{0:T}} P_\theta(a_{0:T} \mid x) R(x,a_{0:T})$ is a baseline that reduces the variance of the gradient estimation without introducing bias. Having a separate network to predict the baseline is an interesting future direction. 

While REINFORCE assumes a stochastic policy, we use beam search for gradient estimation. Thus, in contrast with common practice of approximating the gradient by sampling from the model, we use the top-$k$ action sequences (programs) in the beam with normalized probabilities.
This allows training to focus on sequences with high probability, which are on the decision boundaries, and reduces the variance of the gradient.



Empirically (and in line with prior work), REINFORCE converged slowly and often got stuck in local optima (see Section \ref{sec:experiment}). 
The difficulty of training resulted from the sparse reward signal in the large search space, which caused model probabilities for programs with non-zero reward to be very small at the beginning. 
If the beam size $k$ is small, good programs fall off the beam, leading to zero gradients for all programs in the beam. 
If the beam size $k$ is large, training is very slow, and the normalized probabilities of good programs when the model is untrained are still very small, leading to (1) near zero baselines, thus near zero gradients on ``bad" programs (2)  near zero gradients on good programs due to the low probability  $P_\theta(a_{0:T} \mid x)$. To combat this, we present an alternative training strategy based on maximum-likelihood.



\paragraph{Iterative ML}
If we had gold programs, we could directly optimize their likelihood. Since we do not have gold programs, we can perform an iterative procedure (similar to hard Expectation-Maximization (EM)), where we search for good programs given fixed parameters, and then optimize the probability of the best program found so far. We do decoding on an example with a large beam size and declare $a^{best}_{0:T}(x)$ to be the pseudo-gold program, which achieved highest reward with shortest length among the programs decoded on $x$ in all previous iterations. Then, we can optimize the ML objective:
\begin{equation} \label{eq:ml}
J^{ML}(\theta)= \sum_{x} \log{P_\theta(a^{best}_{0:T}(x) \mid x)}
\end{equation}
A question $x$ is not included if we did not find any program with positive reward. 

Training with iterative ML is fast because there is at most one program per example and the gradient is not weighted by model probability. while decoding with a large beam size is slow, we could train for multiple epochs after each decoding. This iterative process  has a bootstrapping effect that a better model leads to a better program $a^{best}_{0:T}(x)$ through decoding, and a better program $a^{best}_{0:T}(x)$ leads to a better model through training.

Even with a large beam size, some programs are hard to find because of the large search space. A common solution to this problem is to use curriculum learning \cite{zaremba2015reinforcement,reed2015neural}.
The size of the search space is controlled by both the set of functions used in the program and the program length. We apply curriculum learning by gradually increasing both these quantities (see details in Section \ref{sec:experiment}) when performing iterative ML.


Nevertheless, iterative ML uses only pseudo-gold programs and does not directly optimize the objective we truly care about. This has two adverse effects:
(1) The best program $a^{best}_{0:T}(x)$ could be a spurious program that accidentally produces the correct answer (e.g., using the property \textsc{PlaceOfBirth} instead of  \textsc{PlaceOfDeath} when the two places are the same), and thus does not generalize to other questions.
(2) Because training does not observe full negative programs, the model often fails to distinguish between tokens that are related to one another. For example, differentiating \textsc{ParentsOf} vs. \textsc{SiblingsOf} vs. \textsc{ChildrenOf} can be challenging. 
We now present learning where we combine iterative ML with REINFORCE.




\begin{algorithm}[ht!]
\caption{IML-REINFORCE}\label{alg:aug-reinf}
\begin{algorithmic}
\footnotesize{
\State \textbf{Input:} 
question-answer pairs $\mathbb{D}=\{(x_i, y_i)\}$, 
mix ratio $\alpha$, 
reward function $R(\cdot)$, 
training iterations $N_{ML}$, $N_{RL}$, and 
beam sizes $B_{ML}$, $B_{RL}$.
\State \textbf{Procedure:}
\State Initialize $C^*_x=\emptyset$ the best program so far for $x$
\State Initialize model $\theta$ randomly \Comment{Iterative ML} 
\For{ $n=1$ to $N_{ML}$} 
  \For{ $(x,y)$ in $D$} 
    \State $\mathbb{C} \leftarrow$ Decode $B_{ML}$ programs given $x$
    \For{ $j$ in $1 ... |\mathbb{C}|$}
        \If{ $R_{x,y}(C_j) > R_{x,y}(C^*_x)$ } $C^*_x \leftarrow C_j$ \EndIf
    \EndFor
  \EndFor
  \State $\theta \leftarrow$ ML training with $\mathbb{D}_{ML} = \{(x, C^*_x)\}$
\EndFor
\State Initialize model $\theta$ randomly \Comment{REINFORCE}
\For{ $n=1$ to $N_{RL}$} 
  \State $\mathbb{D}_{RL} \leftarrow \emptyset$ is the RL training set  
  \For{ $(x, y)$ in $D$} 
    \State $\mathbb{C} \leftarrow$ Decode $B_{RL}$ programs from $x$
    \For{ $j$ in $1 ... |\mathbb{C}|$}
        \If{ $R_{x,y}(C_j) > R_{x,y}(C^*_x)$ } $C^*_x \leftarrow C_j$ \EndIf
    \EndFor
    \State $\mathbb{C} \leftarrow \mathbb{C} \cup \{C^*_x\}$
    \For{ $j$ in $1 ... |\mathbb{C}|$}
        \State $\hat{p}_j \leftarrow (1 - \alpha) \cdot \frac{p_j}{\sum_{j'} p_{j'}}$ where $p_j=P_\theta(C_j \mid x)$
      \If{ $C_j = C^*_x$ } $\hat{p}_j \leftarrow  \hat{p}_j + \alpha$ \EndIf
      \State $\mathbb{D}_{RL} \leftarrow \mathbb{D}_{RL} \cup \{(x, C_j, \hat{p}_j)\}$
    \EndFor
  \EndFor
  \State $\theta \leftarrow$ REINFORCE training with $\mathbb{D}_{RL}$
\EndFor
}
\end{algorithmic}
\end{algorithm}

\paragraph{Augmented REINFORCE}
To bootstrap REINFORCE, we can use iterative ML to find pseudo-gold programs, and then add these programs to the beam with a reasonably large probability. This is similar to methods from imitation learning \cite{ross2011reduction,jiang2012learned} that define a proposal distribution by linearly interpolating the model distribution and an oracle.


Algorithm~\ref{alg:aug-reinf} describes our overall training procedure. We first run iterative ML for $N_{ML}$ iterations and record the best program found for every example $x_i$.
Then, we run REINFORCE, where we normalize the probabilities of the programs in beam to sum to $(1-\alpha)$ and add $\alpha$ to the probability of the best found program $C^*(x_i)$. 
Consequently, the model always puts a reasonable amount of probability on a program with high reward during training. Note that we randomly initialized the parameters for REINFORCE, since initializing from the final ML parameters seems to get stuck in a local optimum and produced worse results.

On top of imitation learning, our approach is related to the common practice in reinforcement learning \cite{schaul2016prioritized} to replay rare successful experiences to reduce the training variance and improve training efficiency. This is also similar to recent developments \cite{wu2016gnmt} in machine translation, where ML and RL objectives are linearly combined, because anchoring the model to some high-reward outputs stabilizes training.

\makeatletter
\def\BState{\State\hskip-\ALG@thistlm}
\makeatother


\section{Experiments and Analysis} 
\label{sec:experiment}

We now empirically show that NSM can learn a semantic parser from weak supervision over a large KB. We evaluate on \textsc{WebQuestionsSP}, a challenging semantic parsing dataset with strong baselines.
Experiments show that NSM achieves new state-of-the-art performance on \textsc{WebQuestionsSP} with weak supervision, and significantly closes the gap between weak and full supervisions for this task. 

\subsection{The \textsc{WebQuestionsSP} dataset}
\label{sec:webquestions}

%

The \textsc{WebQuestionsSP} dataset \cite{yih2016webquestionssp} 
contains full semantic parses for a subset of the questions from 
\textsc{WebQuestions} \cite{berant2013semantic}, because  18.5\% of the original dataset were found to be ``not answerable". 
It consists of  3,098 question-answer pairs for training and 1,639 for testing, which were collected using Google Suggest API, and the answers were originally obtained using Amazon Mechanical Turk workers. They were updated in \cite{yih2016webquestionssp} by annotators who were familiar with the design of Freebase and added semantic parses.
We further separated out 620 questions from the training set as a validation set.
For query pre-processing we used an in-house named entity linking system to find the entities in a question. The quality of the entity linker is similar to that of \cite{yih2015semantic} at  $94\%$ of the gold root entities being included. 
Similar to \newcite{dong2016language}, we replaced named entity tokens with a special token ``\textit{ENT}". For example, the question ``\emph{who plays meg in family guy}" is changed to ``\emph{who plays ENT in ENT ENT}". 
This helps reduce overfitting, because instead of memorizing the correct program for a specific entity, the model has to focus on other context words in the sentence, which improves generalization.

Following \cite{yih2015semantic} we used the last publicly available snapshot of Freebase \cite{bollacker2008freebase}. 
Since NSM training requires random access to Freebase during decoding, we preprocessed Freebase by removing predicates that are not related to world knowledge (starting with ``\emph{/common/}", ``\emph{/type/}", ``\emph{/freebase/}"),\footnote{We kept ``\emph{/common/topic/notable\_types}".} and removing all text valued predicates, which are rarely the answer. 
Out of all 27K relations, 434 relations are removed during preprocessing.  
This results in a graph that fits in memory with 23K relations, 82M nodes, and 417M edges.

\subsection{Model Details}

For pre-trained word embeddings, we used the 300 dimension GloVe word embeddings trained on 840B tokens \cite{Pennington2014GloveGV}. On the encoder side, we added a projection matrix to transform the embeddings into 50 dimensions. On the decoder side, we used the same GloVe embeddings to construct an embedding for each property using its Freebase id, and also added a projection matrix to transform this embedding to 50 dimensions.
A Freebase id contains three parts: domain, type, and property. For example, the Freebase id for \textsc{ParentsOf} is \emph{``/people/person/parents"}. \emph{``people"} is the domain, \emph{``person"} is the type and \emph{``parents"} is the property. The embedding is constructed by concatenating the average of word embeddings in the domain and type name to the average of word embeddings in the property name.  For example, if the embedding dimension is 300, the embedding dimension for \emph{``/people/person/parents"} will be 600. The first 300 dimensions will be the average of the embeddings for \emph{``people"} and \emph{``person"}, and the second 300 dimensions will be the embedding for \emph{``parents"}.

The dimension of encoder hidden state, decoder hidden state and key embeddings are all 50.
The embeddings for the functions and special tokens (e.g., ``\emph{UNK}", ``\emph{GO}")  are randomly initialized by a truncated normal distribution with mean=0.0 and stddev=0.1. 
All the weight matrices are initialized with a uniform distribution in $[-\frac{\sqrt{3}}{d}, \frac{\sqrt{3}}{d}]$ where $d$ is the input dimension. 
Dropout rate is set to 0.5, and we see a clear tendency for larger dropout rate to produce better performance, indicating overfitting is a major problem for learning. 

\subsection{Training Details}

In iterative ML training, the decoder uses a beam of size $k=100$ to update the pseudo-gold programs and the model is trained for 20 epochs after each decoding step. We use the Adam optimizer \cite{KingmaB14} with initial learning rate 0.001. 
In our experiment, this process usually converges after a few (5-8) iterations.
%

For REINFORCE training, the best hyperparameters are chosen using the validation set. 
We use a beam of size $k=5$ for decoding, and $\alpha$ is set to 0.1.
Because the dataset is small and some relations are only used once in the whole training set, we train the model on the entire training set for 200 iterations with the best hyperparameters. Then we train the model with learning rate decay until convergence. 
Learning rate is decayed as $g_{t} = g_{0} \times \beta ^{\frac{\max(0, t-t_{s})}{m}}$, where $g_{0}=0.001$, $\beta=0.5$ $m=1000$, and $t_{s}$ is the number of training steps at the end of iteration 200. 




Since decoding needs to query the knowledge base (KB) constantly, the speed bottleneck for training is decoding. We address this problem in our implementation by partitioning the dataset, and using multiple decoders in parallel to handle each partition. 
We use 100 decoders, which queries 50 KG servers, and one trainer. The neural network model is implemented in TensorFlow. Since the model is small, we didn't see a significant speedup by using GPU, so all the decoders and the trainer are using CPU only.

Inspired by the staged generation process in \newcite{yih2015semantic}, curriculum learning includes two steps. 
We first run iterative ML for 10 iterations with programs constrained to only use the ``\emph{Hop}" function and the maximum number of expressions is 2. 
Then, we run iterative ML again, but use both ``\emph{Hop}" and ``\emph{Filter}". The maximum number of expressions is 3, and the relations used by ``\emph{Hop}" are restricted to those that appeared in $a^{best}_{0:T}(q)$ in the first step. 

\subsection{Results and discussion}

We evaluate performance using the offical evaluation script for \textsc{WebQuestionsSP}. Because the answer to a question may contain multiple entities or values, precision, recall and F1 are computed based on the output of each individual question, and average F1 is reported as the main evaluation metric. Accuracy measures the proportion of questions that are answered exactly. 

A comparison to STAGG, the previous state-of-the-art model \cite{yih2016webquestionssp,yih2015semantic}, is shown in Table~\ref{tab-result}. Our model beats STAGG  with weak supervision by a significant margin on all metrics, while relying on no feature engineering or hand-crafted rules. When STAGG is trained with strong supervision it obtains an F1 of 71.7, and thus NSM closes half the gap between training with weak and full supervision.

\begin{table}[h!]
\centering
  \begin{tabular}{ l | c | c | c | c }
  	\hline
    Model & Prec. & Rec. & F1 & Acc.  \\ \hline \hline
    \textit{STAGG } & 67.3 & 73.1 & 66.8 & 58.8 \\ 
    \textit{NSM } & 70.8 & 76.0 & \textbf{69.0} & 59.5 \\
    \hline
  \end{tabular}
\caption{Results on the test set. Average F1 is the main evaluation metric and NSM outperforms STAGG with no domain-specific knowledge or feature engineering. 
}
\label{tab-result}
\end{table}


Four key ingredients lead to the final performance of NSM. The first one is the neural computer interface that provides code assistance by checking for syntax and semantic errors.
We find that semantic checks are very effective for open-domain KBs with a large number of properties.
For our task, the average number of choices is reduced from 23K per step (all properties) to less than 100 (the average number of properties connected to an entity). 

The second ingredient is augmented REINFORCE training. Table~\ref{tab-rl-ml} compares augmented REINFORCE, REINFORCE, and iterative ML on the validation set.
REINFORCE gets stuck in local optimum and performs poorly. Iterative ML training is not directly optimizing the F1 measure, and achieves sub-optimal results.
In contrast, augmented REINFORCE is able to bootstrap using pseudo-gold programs found by iterative ML and achieves the best performance on both the training and validation set. 

\begin{table}[h!]
\centering
  \begin{tabular}{ l | c | c  }
  	\hline
    Settings & Train F1 & Valid F1   \\ \hline \hline
    \textit{Iterative ML} & 68.6 & 60.1 \\ 
    \textit{REINFORCE} & 55.1 & 47.8 \\
    \textit{Augmented REINFORCE} & 83.0 & \textbf{67.2} \\
    \hline
    \hline
  \end{tabular}
\caption{Average F1 on the validation set for augmented REINFORCE, REINFORCE, and iterative ML.}
\label{tab-rl-ml}
\end{table}

The third ingredient is curriculum learning during iterative ML.
We compare the performance of the best programs found with and without curriculum learning in Table~\ref{tab-cur}. We find that the best programs found with curriculum learning are substantially better than those found without curriculum learning by a large margin on every metric. 

\begin{table}[h!]
\centering
  \begin{tabular}{ l | c | c | c | c }
  	\hline
    Settings & Prec.& Rec.& F1 & Acc.  \\ \hline \hline
    \textit{No curriculum} & 79.1 & 91.1 & 78.5 & 67.2 \\ 
    \textit{Curriculum} & 88.6 & 96.1 & 89.5 & 79.8 \\
    \hline
    \hline
  \end{tabular}
\caption{Evaluation of the programs with the highest F1 score in the beam ($a_{0:t}^{best}$) with and without curriculum learning.
}
\label{tab-cur}
\end{table}

The last important ingredient is reducing overfitting. Given the small size of the dataset, overfitting is a major problem for training neural network models. We show the contributions of different techniques for controlling overfitting in Table \ref{tab-ablation}. Note that after all the techniques have been applied, the model is still overfitting with 
training F1@1=83.0\% and validation F1@1=67.2\%.

\begin{table}[h!]
\centering
  \begin{tabular}{ l | c }
  	\hline
    Settings & $\Delta$ F1@1   \\ \hline \hline
    \textit{$-$Pretrained word embeddings} & $-5.5$ \\ 
    \textit{$-$Pretrained property embeddings} & $-2.7$ \\ 
    \textit{$-$Dropout on GRU input and output} & $-2.4$ \\
    \textit{$-$Dropout on softmax} & $-1.1$ \\
    \textit{$-$Anonymize entity tokens} & $-2.0$ \\
    \hline
    \hline
  \end{tabular}
\caption{Contributions of different overfitting techniques on the validation set.}
\label{tab-ablation}
\end{table}

\begin{table}[h!]
\centering
  \begin{tabular}{ l | c | c | c | c }
  	\hline
    \#Expressions & 0 & 1 & 2 & 3 \\ \hline \hline
    \textit{Percentage} & 0.4\% & 62.9\% & 29.8\% & 6.9\% \\ 
    \textit{F1} & 0.0 & 73.5 & 59.9 & 70.3  \\
    \hline
    \hline
  \end{tabular}
\caption{Percentage and performance of model generated programs with different complexity (number of expressions).}
\label{tab-complexity}
\end{table}

Among the programs generated by the model, a significant portion (36.7\%) uses more than one expression. From Table~\ref{tab-complexity}, we can see that the performance doesn't decrease much as the compositional depth increases, indicating that the model is effective at capturing compositionality. 
We observe that programs with three expressions use a more limited set of properties, mainly focusing on answering a few types of questions such as ``who plays meg in family guy'', ``what college did jeff corwin go to'' and ``which countries does russia border''.
In contrast,  programs with two expressions use a more diverse set of properties, which could explain the lower performance compared to programs with three expressions.

\paragraph{Error analysis}
Error analysis on the validation set shows two main sources of errors:
\begin{enumerate}[noitemsep,topsep=0pt]
\item \textbf{Search failure}: Programs with high reward are not found during search for pseudo-gold programs, either because the beam size is not large enough, or because the set of functions implemented by the interpreter is insufficient.  
The 89.5\% F1 score in Table~\ref{tab-cur} indicates that at least $10\%$ of the questions are of this kind.
\item \textbf{Ranking failure}: Programs with high reward exist in the beam, but are not ranked at the top during decoding. 
Because the training error is low, this is largely due to overfitting or spurious programs. 
The 67.2\% F1 score in Table~\ref{tab-rl-ml} indicates that about $20\%$ of the questions are of this kind.
\end{enumerate}

\section{Related work}
Among deep learning models for program induction, 
Reinforcement Learning Neural Turing Machines (RL-NTMs) \cite{zaremba2015reinforcement} are the most similar to NSM, as a non-differentiable machine is controlled by a sequence model. Therefore, both models rely on REINFORCE for training.
The main difference between the two is the abstraction level of the programming language. 
RL-NTM uses lower level operations such as memory address manipulation and byte reading/writing, while
NSM uses a high level programming language over a large knowledge base that includes operations such as following properties from entities, or sorting  based on a property, which is more suitable for representing semantics.
Earlier works such as OOPS \cite{Schmidhuber04} has desirable characteristics, for example, the ability to define new functions.
These remain to be future improvements for NSM.

We formulate NSM training as an instance of 
reinforcement learning \cite{sutton1998reinforcement} in order
to directly optimize the task reward of the structured prediction problem \cite{norouzi2016,li2016deep,Yu2015Skim}.
Compared to imitation learning methods~\cite{daume09searn, ross2011reduction} that interpolate a model distribution with an oracle, NSM needs to solve a challenging search problem of training from weak supervisions in a large search space.
Our solution employs two techniques (a) a symbolic ``computer" helps find good programs by pruning the search space (b) an iterative ML training process, where beam search is used to find pseudo-gold programs.
Wiseman and Rush \cite{wiseman2016beam} proposed a max-margin approach to train a sequence-to-sequence scorer. However, their training procedure is more involved, and we did not implement it in this work. 
MIXER \cite{ranzato2015sequence} also proposed to combine ML training and REINFORCE, but they only considered tasks with full supervisions. 
Berant and Liang \cite{berant2015imitation} applied imitation learning to semantic parsing, but still requires hand crafted grammars and features.

NSM is similar to Neural Programmer \cite{Neelakantan2015NeuralPI} and Dynamic Neural Module Network  \cite{andreas2016compose}  in that they all solve the problem of semantic parsing from structured data, and generate programs using similar semantics. 
The main difference between these approaches is how an intermediate result (the memory) is represented. 
Neural Programmer and Dynamic-NMN chose to represent results as vectors of weights (row selectors and attention vectors), which enables backpropagation and search through all possible programs in parallel. However, their strategy is not applicable to a large KB such as Freebase, which contains about 100M entities, and more than 20k properties.
Instead, NSM chooses a more scalable approach, where the ``computer" saves intermediate results, and the neural network only refers to them with variable names (e.g., ``$R_1$" for all cities in the US). 



NSM is similar to the Path Ranking Algorithm (PRA) \cite{lao2011random} in that semantics is encoded as a sequence of actions, and denotations are used to prune the search space during learning. NSM is more powerful than PRA by 1) allowing more complex semantics to be composed through the use of a key-variable memory; 2) controlling the search procedure with a trained neural network, while PRA only samples actions uniformly; 3) allowing input questions to express complex relations, and then dynamically generating action sequences. PRA can combine multiple semantic representations to produce the final prediction, which remains to be future work for NSM.

\section{Conclusion}

We propose the 
Manager-Programmer-Computer framework for neural program induction. 
It integrates neural networks with a symbolic \textit{non-differentiable} computer to support \textit{abstract}, \textit{scalable} and \textit{precise} operations through a friendly \textit{neural computer interface}. 
Within this framework, we introduce the Neural Symbolic Machine, which integrates a neural sequence-to-sequence ``programmer" with key-variable memory, and a symbolic Lisp interpreter with code assistance. 
Because the interpreter is non-differentiable and to directly optimize the task reward, we apply REINFORCE and use pseudo-gold programs found by an iterative ML training process to bootstrap training. NSM achieves new state-of-the-art results on a challenging semantic parsing dataset with weak supervision, and significantly closes the gap between weak and full supervision.
It is trained end-to-end, and does not require any feature engineering or domain-specific knowledge.   

\subsubsection*{Acknowledgements}

We thank for discussions and help from Arvind Neelakantan, Mohammad Norouzi, Tom Kwiatkowski, Eugene Brevdo, Lukasz Kaizer, Thomas Strohmann, Yonghui Wu, Zhifeng Chen, Alexandre Lacoste, and John Blitzer.
The second author is partially supported by the Israel Science Foundation, grant 942/16.



\bibliography{acl2017}
\bibliographystyle{acl_natbib}

\onecolumn
\appendix
\section{Supplementary Material}
\label{sec:supplemental}
\subsection{Extra Figures}

\begin{figure}[h]
\centering
\includegraphics[width=\textwidth]{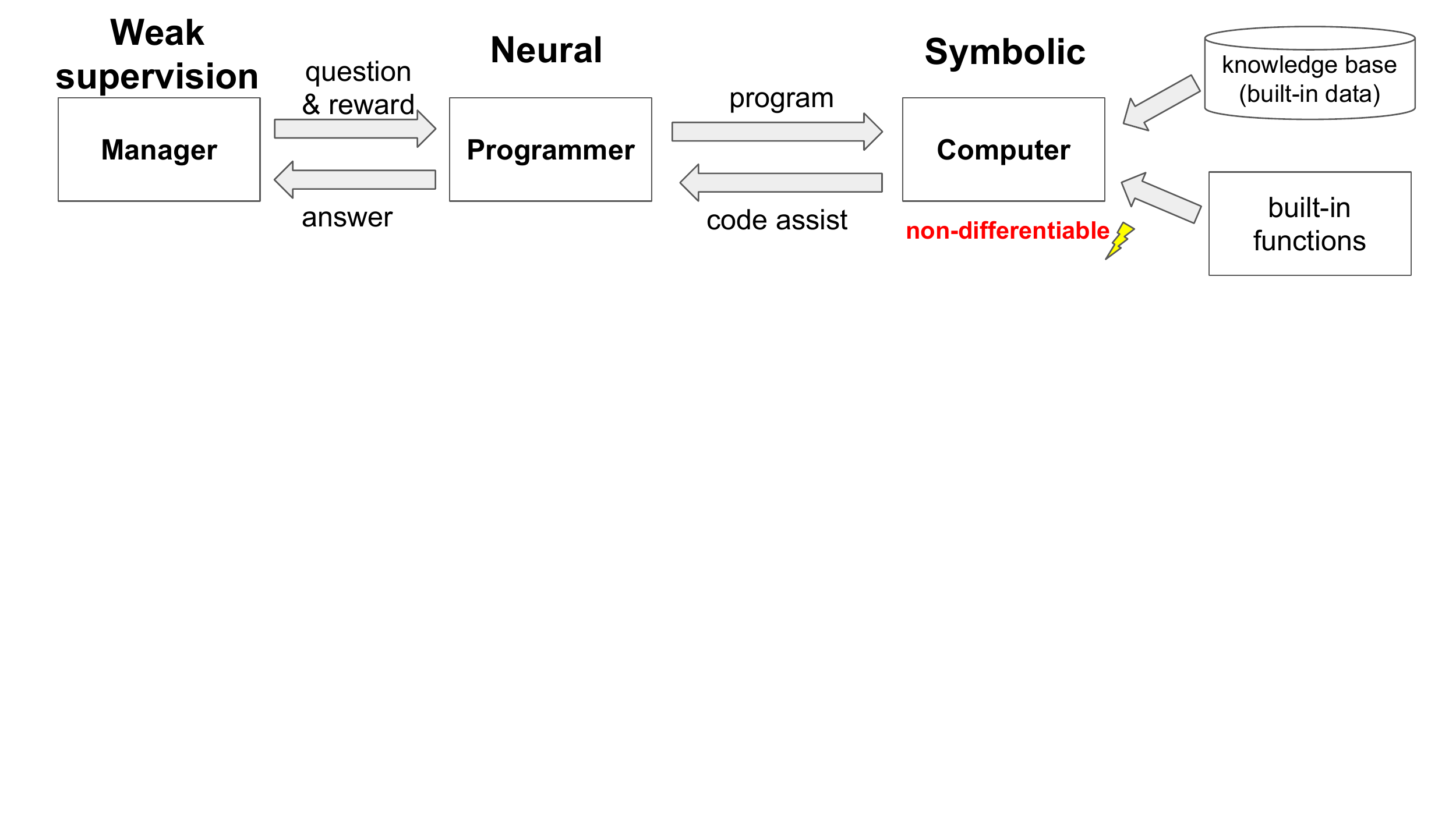}
\vspace{-2.3in}
\caption{\label{fig:bpc} Manager-Programmer-Computer framework }
\end{figure}

\begin{figure}[h!]
\centering
\includegraphics[width=\textwidth]{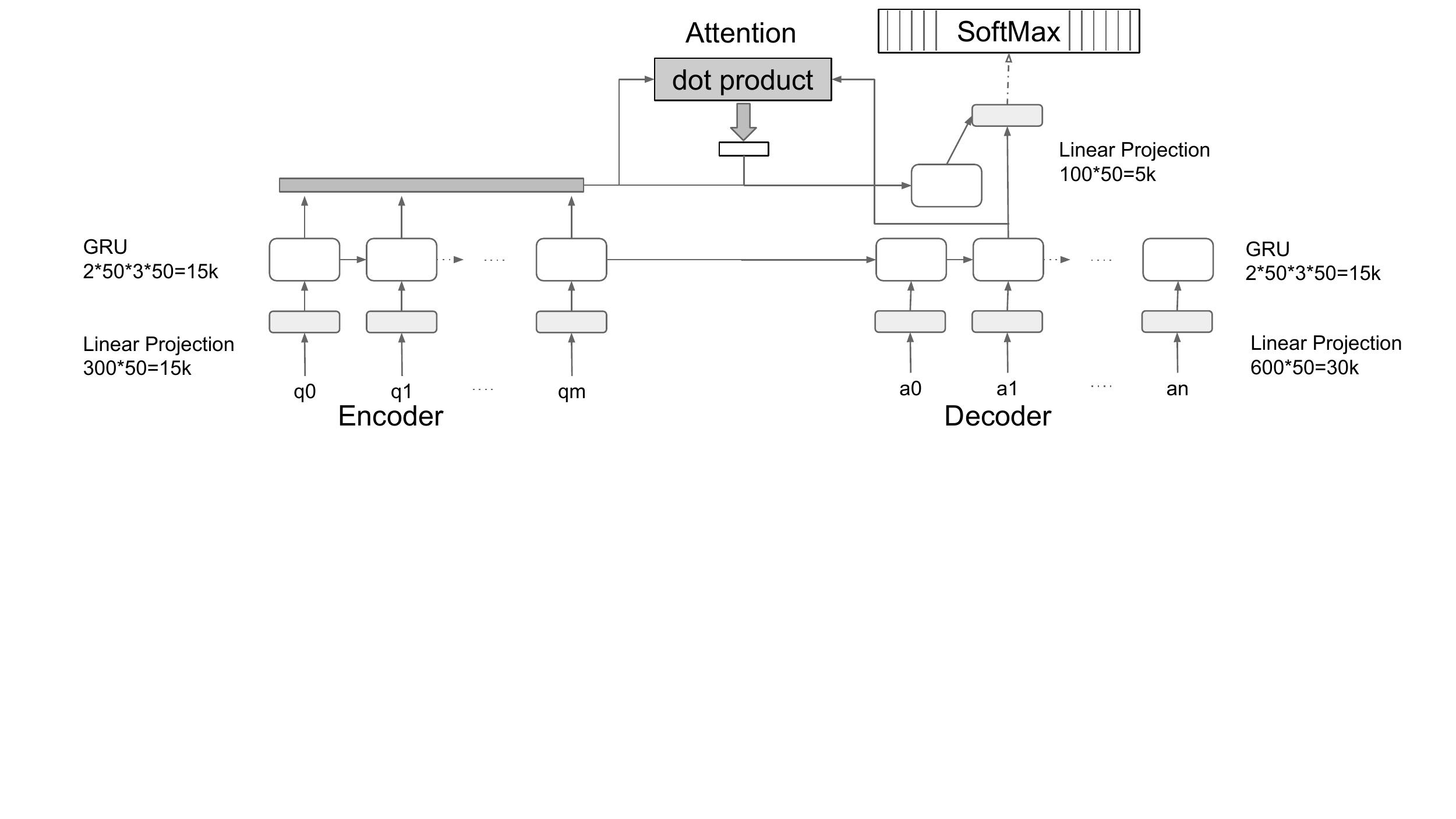}
\vspace{-1.7in}
\caption{\label{fig:model} Seq2seq model architecture with dot-product attention and dropout at GRU input, output, and softmax layers.
}
\end{figure}

\begin{figure}[h!]
\centering
\vspace{-0.1in}
\includegraphics[width=0.7\textwidth]{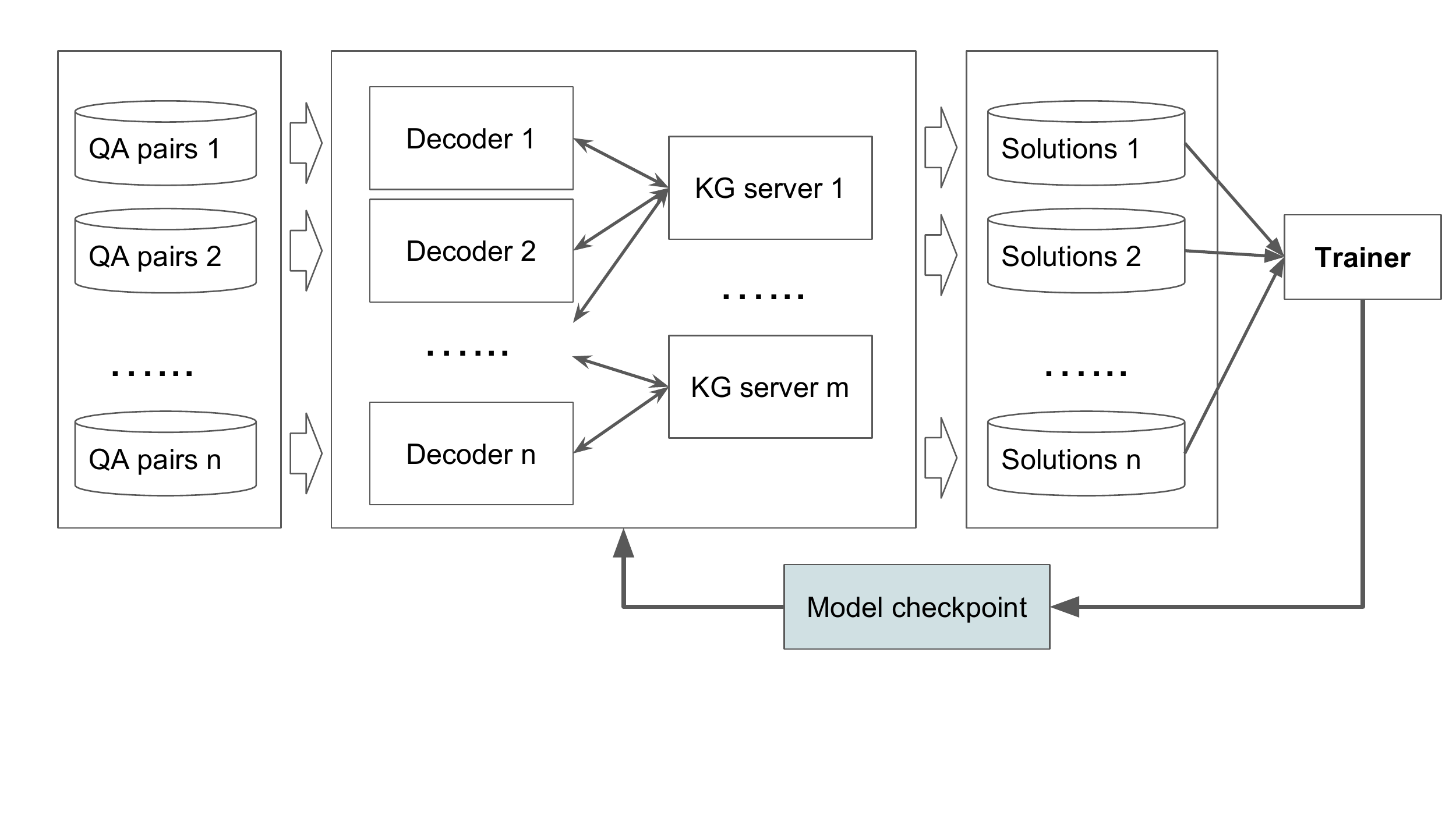}
\vspace{-0.5in}
\caption{\label{fig:architecture} System Architecture. 
100 decoders, 50 KB servers and 1 trainer.}
\end{figure}

\end{document}